\documentclass[letterpaper]{article}
\usepackage[preprint]{aaai2027}
\usepackage[hyphens]{url}
\usepackage{graphicx}
\usepackage{natbib}
\usepackage{caption}
\usepackage{booktabs}
\usepackage[table]{xcolor}
\usepackage{amsmath}
\usepackage{amssymb}
\usepackage{algorithm}
\usepackage{algorithmic}
\usepackage{newfloat}
\usepackage{listings}
\usepackage{subcaption}
\DeclareCaptionStyle{ruled}{labelfont=normalfont,labelsep=colon,strut=off}
\floatstyle{ruled}
\newfloat{listing}{tb}{lst}{}
\floatname{listing}{Listing}

\title{MRC-MMAgent: Reinforcement Learning for Multi-Role Collaboration in Multimodal Tool-Using Agents
}
\author{
Shengqin Wang\equalcontrib,
Jie Jin\equalcontrib,
Xiaofan Li,
Wentao Yan,
Yu Cheng, 
Qingsong Hu
Yihang Chen, 
Weilin Luo, 
Jingyu Gong, 
Xin Tan, 
Yuan Xie,
Zhizhong Zhang 
}

\affiliations{
East China Normal University,
Shanghai Innovation Institute,
University College London,
Huawei Noah's Ark Lab
}

\begin{document}

\maketitle



\begin{abstract}
Multimodal tool-using agents augment vision-language models with access to external information and capabilities, enabling them to solve knowledge-intensive and visually complex tasks that cannot be addressed using visual context alone. However, existing reinforcement learning pipelines largely train single-role agents to search and answer using terminal rewards, overlooking how humans inspect retrieved evidence, reformulate queries, and revise conclusions during search. Consequently, they fail to optimize this self-corrective process over complete interaction trajectories. We propose MRC-MMAgent, a multi-role collaborative reinforcement learning framework that incorporates trajectory assessment and continuation directly into the executable agent loop. First, the framework establishes a multi-role collaborative loop that integrates solving, inspection, and continuation, enabling the agent to acquire evidence through tools, evaluate the preceding trajectory, and resume problem solving when the current answer is insufficiently supported. Second, we introduce a role-coupled learning objective that combines task success,
interaction validity, and successful correction with full-trajectory policy optimization across the decisions generated by all roles. Extensive experiments show that, using only the original baseline training set without constructing additional datasets, our method consistently outperforms the corresponding base models and achieves leading or highly competitive performance across benchmarks.
\end{abstract}

\section{Introduction}


Vision-Language Models (VLMs)~\cite{liu2024improved,chen2024internvl,liu2023visual,zhu2025internvl3,zhang2024mathverse} have rapidly improved on perception, recognition, and visual question answering, yet many user questions cannot be answered from images alone. Multimodal agents address this limitation by coupling a foundation model with tools such as reverse image search, text search, region zooming, and retrieval~\citep{chen2022murag,chen2024mllm,yu2024rankrag, hu2023avis,jiang2024mmsearch,zhang2024vision}. Systems such as MMSearch-R1, WebWatcher, DeepEyes2 optimize tool-augmented policies for multimodal search and visual question answering ~\citep{wu2025mmsearch,geng2025webwatcher,hong2025deepeyesv2}. SenseNova-MARS~\citep{chng2025sensenova} further constructs high-resolution multimodal agent data and proposes BN-GSPO for training autonomous tool-using VLMs. These methods establish Reinforcement Learning (RL) as a foundation for developing agents’ core tool-use capabilities.

\begin{figure}[t]
\centering
\includegraphics[width=0.45\textwidth]{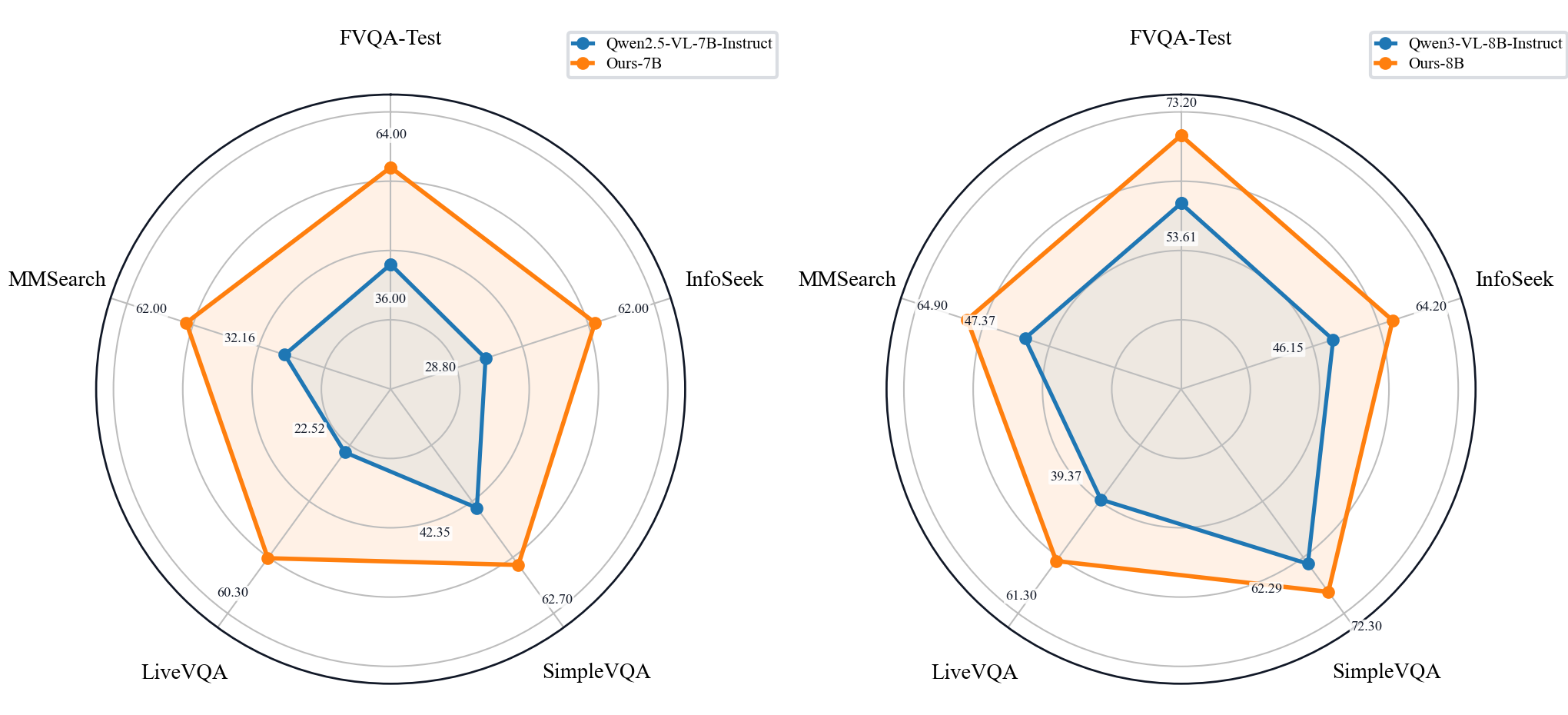}
\caption{Performance comparison of agents across different scales. Each radar chart reports results on the same five evaluation datasets.}
\label{fig:radar_comparison}
\end{figure}
\begin{figure*}[t]
\centering
\includegraphics[width=0.98\textwidth]{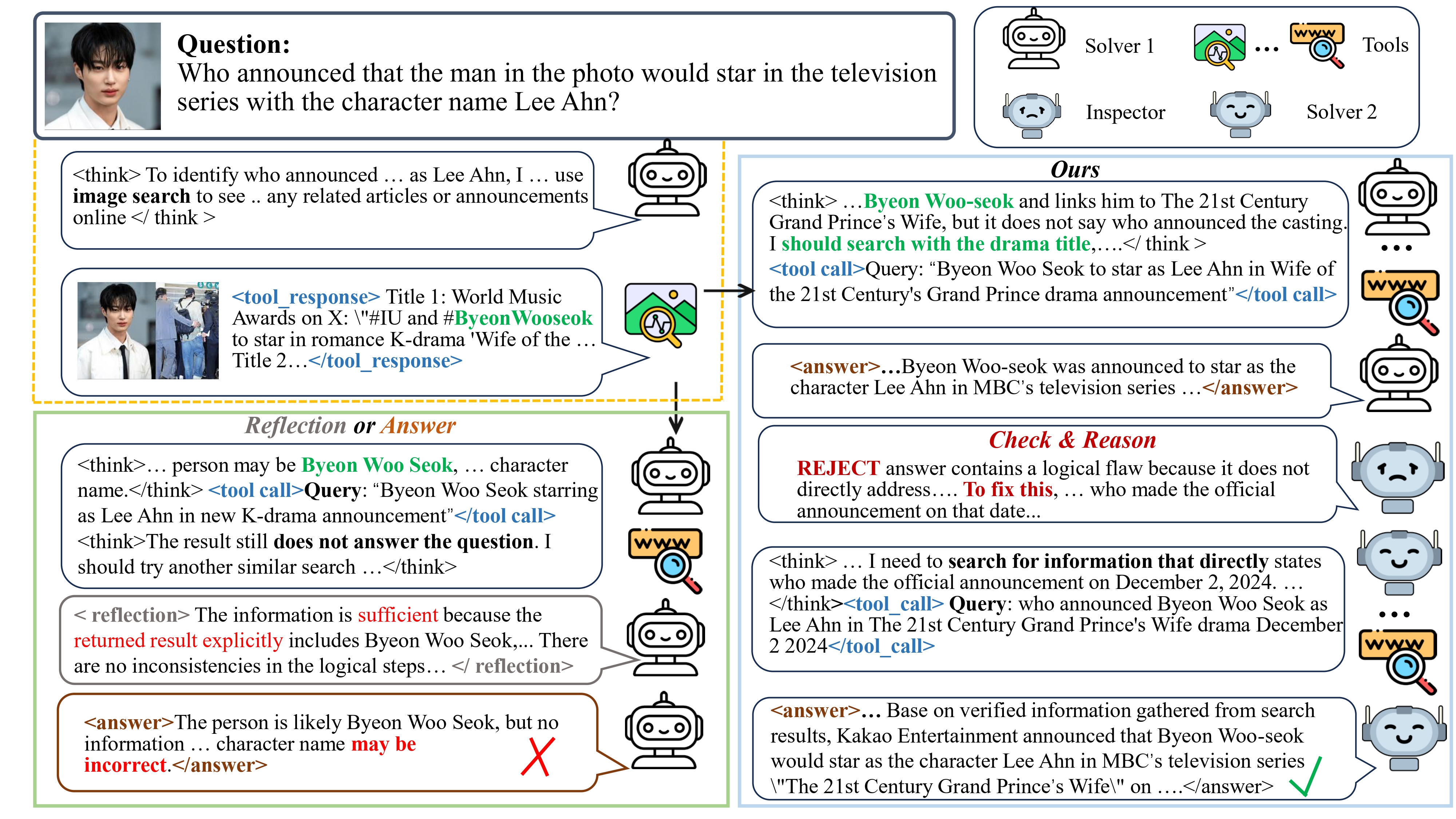}
\caption{Case Study. The left panel shows a failure case where the current agent loop produces an incorrect answer. The right panel shows that after the inspector reviews the visual-textual trajectory, it detects the failure; the solver then resumes from the accumulated context and successfully corrects the answer.}
\label{fig:3}
\end{figure*}

However, prevailing RL formulations train multimodal tool agents as
single-role solvers under terminal-answer supervision, leaving a critical
control process weakly supervised. This formulation has two limitations: its single-role design lacks a dedicated mechanism for evidence assessment and continuation control, while terminal-only rewards fail to jointly optimize assessment, control, and revision across the full trajectory. First, as illustrated in Figure~\ref{fig:3}, a single agent loop may retrieve relevant snippets and still answer based on the wrong comparison basis, they lack an explicit, trainable mechanism that connects evidence assessment
with trajectory control, making it difficult to identify under-supported
answers and initiate targeted continuation. Second, as shown in the "reflection" part of Figure~\ref{fig:3}, where we insert the inspection stage into the solver's loop, the reflection merely repeats or rationalizes the initial reasoning. Mirroring human information seeking, the agent assesses the preceding solution trajectory, finalizes the answer when it is sufficiently supported, and otherwise resumes tool use to refine its reasoning and revise the answer. This conditional assessment-and-continuation process closes the loop over the entire interaction, highlighting the importance of full-trajectory optimization for reliable reasoning.

To address these limitations, we propose a multi-role collaborative
reinforcement learning framework for multimodal tool agents. First, we organize
solving, inspection, and continuation as role-specific behaviors within a
unified executable loop. The solver gathers external evidence and proposes a
candidate answer, while the inspector assesses whether the accumulated
visual--textual observations support it. If the evidence is sufficient, the
rollout terminates; otherwise, the Inspector returns control to the solver for
targeted tool use and revision. This design turns evidence assessment into a
trainable control decision and enables all roles to co-adapt within a shared
policy. Second, we formulate a role-coupled learning objective. We define behavioral rewards tailored to solving, inspection, and continuation, and aggregate them into a trajectory-level reward. Full-trajectory GRPO applies the aggregated reward signal to all role decisions rather than confining each role-specific signal to its corresponding role, thereby coupling optimization across the complete trajectory. Theoretical analysis and controlled ablations validate both the reward and optimization designs. Using only the MMSearch-R1 training set, with no additional supervised fine-tuning (SFT) trajectories or RL data, our method improves upon the base model by 11.1 points. Extensive experiments show that, without constructing any extra datasets, our method consistently outperforms the corresponding base models and achieves leading or highly competitive performance across search-intensive and visually complex benchmarks. Our contributions are:
\begin{itemize}
    \item We propose a multi-role collaborative reinforcement learning
    framework that organizes solving, trajectory inspection, and continuation
    within a unified executable loop, enabling evidence-conditioned
    finalization and targeted recovery in multimodal tool use.

    \item We formulate a role-coupled learning objective that combines a
    trajectory reward for task success, interaction validity, calibrated
    finalization, and effective correction with full-trajectory  over all
    generated role decisions. 
    \item Across diverse multimodal benchmarks, our method delivers consistent gains over
its base models and leading or competitive results among same-scale systems, without auxiliary SFT trajectories or additional training data.
\end{itemize}
\section{Related Work}

\subsection{Multimodal Tool-Using Agents}
Recent multimodal search and visual question answering agents~\citep{wu2025mmsearch,chng2025sensenova,hong2025deepeyesv2,wang2025pixel} continuously acquire problem-solving capabilities by training VLMs to use tools. However, most current agents rely on expensive SFT trajectories to regularize model behavior, while improving performance through terminal rewards alone. This fixed behavioral regime neglects effective evaluation of the model's reasoning process and further limits the ability to perform meaningful revision.

\subsection{Multi-Agent and Multi-Module Tool-Use RL}
Multi-agent frameworks optimize role decomposition among planners, workers, and specialized agents~\cite{li2025flow,mo2025multi,park2025maporl,chen2026videochat}, while other works improve process supervision and multimodal tool use~\cite{tan2025process,zhang2026vsearcher,dong2025tool,zhao2026pyvision,hong2025deepeyesv2}. Reflection methods such as Look-Back and ReGRPO revisit evidence or use teacher feedback to revise solutions~\citep{yang2026look,klissarov2026improving,zhang2026regrpo}, but do not jointly learn reference-free inspection and recovery within a multimodal tool loop (see Supp. A for detailed comparisons). In contrast, MRC-MMAgent trains solver, inspector, and continuation within one shared policy, without auxiliary SFT or teacher critiques, using a role-coupled reward and full-trajectory updates to jointly optimize calibrated finalization and successful correction.

\begin{figure*}[htb]
\centering
\includegraphics[width=0.92\textwidth]{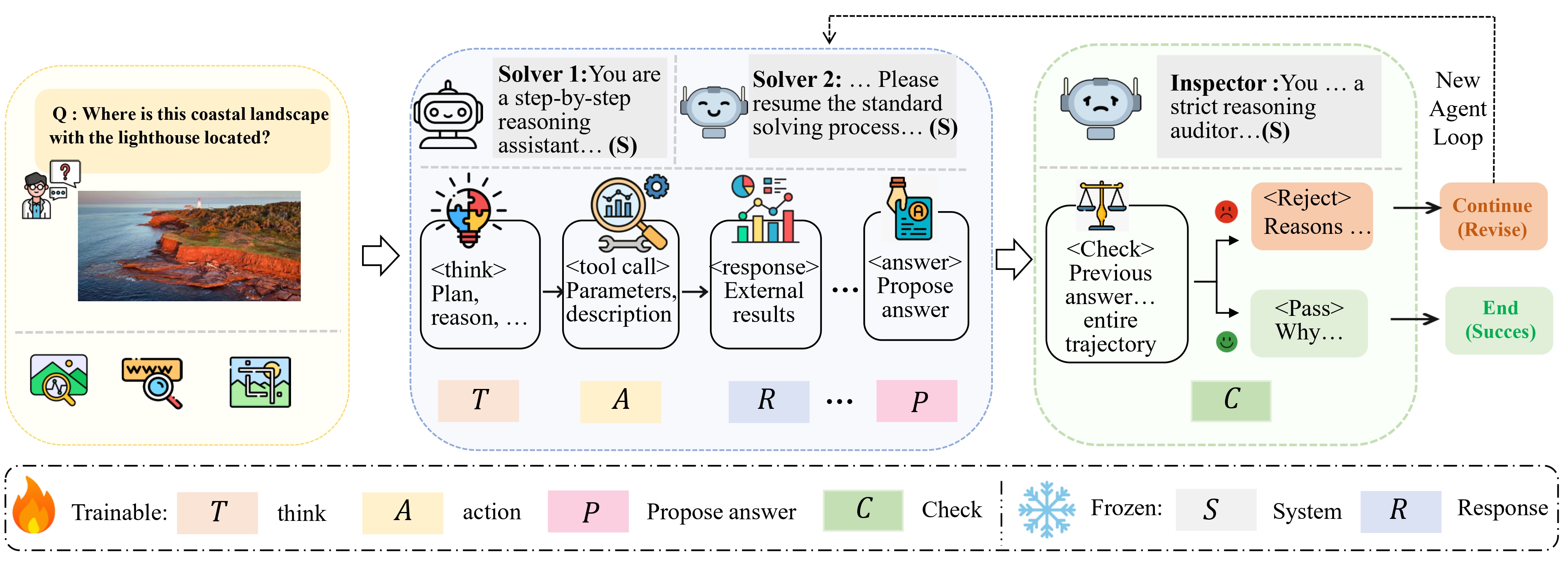}
\caption{Overview of the proposed multi-role collaborative training framework for multimodal tasks. The Solver iteratively tackles the initial query, while the Inspector analyzes visual-textual trajectories to generate suggestions. Failed agent loops are re-executed based on historical trajectories to ensure convergence.}
\label{fig:2}
\end{figure*}
\section{Method}

\subsection{Overview}

Figure~\ref{fig:2} presents our multi-role collaborative reinforcement learning framework, which couples an executable solve--inspect--continue loop with trajectory-level learning. A shared policy $\pi_\theta$ is conditioned by frozen prompts $S_1$, $S_3$, and $S_2$ to act as the initial solver, inspector, and continuation solver, respectively. The solver gathers evidence and proposes an answer; without access to the reference answer, correctness labels, or future outcomes, the inspector determines whether the trajectory sufficiently supports it and either finalizes or requests targeted continuation. We train all roles with a role-coupled reward for task success, interaction validity, and inspection-guided correction. Full-trajectory GRPO then propagates the group-relative signal to every model-generated decision, allowing the shared policy to co-adapt across the complete interaction.

\begin{table*}[ht]
\centering
\setlength{\tabcolsep}{6pt}
\begin{tabular}{lcccccc}
\toprule
Method & Avg. & FVQA-Test & InfoSeek & SimpleVQA & LiveVQA & MMSearch \\
\midrule

\rowcolor{gray!10}
\multicolumn{7}{c}{\textit{Direct Answer}} \\
Qwen2.5-VL-7B-Instruct  & 26.7 & 26.3 & 32.0 & 47.9 & 19.6 & 7.6  \\
Qwen3-VL-8B-Instruct    & 25.0 & 24.2 & 23.2 & 42.9 & 23.2 & 11.7 \\
Qwen3-VL-32B-Instruct   & 31.1 & 32.2 & 29.0 & 45.9 & 31.6 & 17.0 \\
GPT-5                   & 50.0 & 54.4 & 54.2 & 61.7 & 44.4 & 35.1 \\
Claude-4-Sonnet         & 40.1 & 35.3 & 57.3 & 50.9 & 38.5 & 18.7 \\
Gemini-3-Flash          & 54.2 & 56.5 & 54.9 & 63.6 & 38.9 & 57.3 \\
Gemini-3-Pro            & 56.5 & 59.2 & 56.3 & 64.1 & 40.1 & 62.6 \\

\midrule
\rowcolor{gray!10}
\multicolumn{7}{c}{\textit{Agentic Model (zero-shot)}} \\
Qwen2.5-VL-7B-Instruct  & 32.4 & 36.0 & 28.8 & 42.4 & 22.5 & 32.2 \\
Qwen3-VL-8B-Instruct    & 49.8 & 53.6 & 46.2 & 62.3 & 39.4 & 47.4 \\
Qwen2.5-VL-32B-Instruct & 51.9 & 52.2 & 50.1 & 65.2 & 42.2 & 49.7 \\
Qwen3-VL-32B-Instruct   & 52.1 & 54.3 & 49.9 & 64.2 & 42.9 & 49.1 \\
GPT-4o                  & 55.8 & 66.3 & 59.6 & 63.7 & 40.1 & 49.1 \\
Gemini-2.5-Flash        & 58.2 & 61.7 & 53.7 & 68.8 & 47.8 & 59.1 \\
GPT-5                   & 59.6 & 62.6 & 56.0 & 70.6 & 56.0 & 52.6 \\
Gemini-3-Flash          & 60.9 & 64.9 & 61.1 & 67.9 & 48.1 & 62.6 \\
Gemini-3-Pro            & 69.8 & 72.6 & 66.5 & 75.9 & 59.7 & 74.3 \\

\midrule
\rowcolor{gray!10}
\multicolumn{7}{c}{\textit{Trained Multimodal Tool Agents}} \\
MMSearch-R1-7B          & 51.1 & 58.0 & 49.0 & 56.4 & 48.4 & 43.9 \\
WebWatcher-7B           & -    & -    & -    & 54.3 & 51.2 & 49.1 \\
DeepEyes2-7B            & -    & 60.6 & 51.1 & 59.4 & -    & \textbf{63.7} \\
MM-DeepResearch-7B      & 60.8 & 61.9 & 58.7 & 62.0 & 60.0 & 61.4 \\
\rowcolor{gray!10}
\textbf{MRC-MMAgent-7B} & \textbf{62.2} & \textbf{64.0} & \textbf{62.0}
                        & \textbf{62.7} & \textbf{60.3} & 62.0 \\

\midrule
SenseNova-MARS-8B       & 64.6 & 67.1 & 61.7 & 70.2 & 56.2 & \textbf{67.8} \\
OpenSearch-VL-8B        & 67.5 & 71.5 & \textbf{70.2} & 71.6 & 59.6 & 64.5 \\
\rowcolor{gray!10}
\textbf{MRC-MMAgent-8B} & \textbf{68.1} & \textbf{74.2} & 65.6
                        & \textbf{74.3} & \textbf{61.3} & 64.9 \\
\bottomrule
\end{tabular}
\caption{Accuracy comparison on five search-intensive benchmarks. Unlike competing trained agents that use additionally constructed
training data, MRC-MMAgent is trained solely on the original baseline (MMSearch-R1)
training set.}
\label{tab:main_results}
\end{table*}
\subsection{Multi-Role Agent Loop}
Each example is a multimodal query $q=(x,u)$, where $x$ denotes one or more images and $u$ is a natural-language question. At loop index $k$, the solver generates a reasoning--action--answer segment:
\begin{equation}
X_k=(T_k,A_k,R_k,...,P_k)
\sim
\pi_\theta(\cdot\mid h_k,S_{r_k}),
\end{equation}
\begin{equation}
S_{r_k}=
\begin{cases}
S_1, & k=0,\\
S_2, & k>0,
\end{cases}
\end{equation}
where the environment executes actions in $A_k$ and appends their observations to history. The inspector then generates:
\begin{equation}
C_k\sim\pi_\theta(\cdot\mid h_k',S_3),
\end{equation}
where $h_k'$ contains the multimodal query, preceding solving trajectory, retrieved visual--textual observations, and candidate answer $P_k$, etc. In particular,
\begin{equation}\label{eq:inspector_information}
(y^\star,\quad \operatorname{Acc}(P_k),\quad y_f)
\notin h_k',
\end{equation}
where $y^\star$ denotes the reference answer. Thus, the inspector must assess sufficiency from the trajectory rather than compare $P_k$ with a known answer. Its output $C_k$ contains a rationale and decision $d_k\in\{\textsc{finalize},\textsc{continue}\}$, verifying evidential support and identifying missing, conflicting, or inadequately grounded evidence. Finalization returns $y_f=P_k$; continuation converts the identified gap into a frozen bridge message $B_k$ and updates:
\begin{equation}
h_{k+1}=h_k'\oplus C_k\oplus B_k,
\end{equation}
after which the solver resumes under $S_2$ with the existing evidence, rejected candidate, and inspection feedback. The loop ends when the answer is accepted or after at most $N_{m}$ continuations. We denote the initial candidate by $y_0=P_0$ and the returned answer by $y_f$.

A complete rollout can therefore be represented compactly as:
\begin{equation}
\tau=
\bigl(q;
S_1,X_0;[S_3,C_k,S_2,X_k]_{k=1}^{K}\bigr),
\end{equation}
where $ K\leq N_{m},$ $R_k$ collects the frozen tool observations in loop $k$. The trainable events are:
\begin{equation}
\mathcal{M}(\tau)
=
\bigcup_{k=0}^{K}\{T_k,A_k,P_k,C_k\},
\end{equation}
\begin{equation}
\mathcal{E}(\tau)=\tau\setminus\mathcal{M}(\tau).
\end{equation}
They decompose by role as:
\begin{equation}\label{eq:role_token_decomposition}
\mathcal{M}(\tau)
=
\mathcal{M}_{S_1}(\tau)
\cup
\mathcal{M}_{I}(\tau)
\cup
\mathcal{M}_{S_2}(\tau),
\end{equation}

where the subsets contain initial-solver, inspector, and
continuation-solver tokens. The context set $\mathcal{E}(\tau)$
contains role prompts, tool observations, and bridge messages.
These events condition subsequent generation but are not sampled
from $\pi_\theta$. Assuming that the environment dynamics are
independent of $\theta$, the trajectory log-likelihood gradient is
\begin{equation}\label{eq:trajectory_likelihood}
\nabla_\theta \log p_\theta(\tau\mid q)
=
\sum_{t\in\mathcal{M}(\tau)}
\nabla_\theta
\log \pi_\theta(a_t\mid h_t).
\end{equation}
Thus, all policy-generated decisions from solving, inspection, and
continuation contribute to the same trajectory-level update,
providing the basis for full-trajectory credit assignment.
Supplementary Material~B gives the complete rollout algorithm.

\subsection{Role-Coupled Trajectory Reward}
Each rollout receives a reward for final-answer accuracy $R_{\mathrm{acc}}$, interaction validity $R_{\mathrm{fmt}}$, and inspection-guided correction $R_{\mathrm{corr}}$. The task evaluator sets $R_{\mathrm{acc}}\in\{0,1\}$ from $y_f$. The format term $R_{\mathrm{fmt}}\in\{0,1\}$ requires valid tool calls, a candidate answer before inspection, a coherent control decision, and well-formed continuation, preventing malformed interactions from being compensated by answer reward.

The correction reward couples the inspector's evidence-conditioned decision with the solver trajectories before and after inspection. Let $I_{\mathrm{cont}}(\tau)$ indicate whether the first inspection triggers continuation, and let $\operatorname{Acc}(y)\in\{0,1\}$ be the environment's post-hoc task evaluator. Neither the reference answer nor the values of $\operatorname{Acc}(y_0)$ and $\operatorname{Acc}(y_f)$ are exposed to the inspector. The Inspector thus assesses evidential sufficiency without answer labels, while its finalization and continuation decisions are optimized from post-hoc trajectory outcomes. We define:
\begin{equation}\label{eq:correction_reward}
R_{\mathrm{corr}}(\tau)=
\begin{cases}
0, & I_{\mathrm{cont}} \wedge \operatorname{Acc}(y_0)=1,\\
1, & I_{\mathrm{cont}} \wedge \operatorname{Acc}(y_0)=0
     \wedge \operatorname{Acc}(y_f)=1,\\
0.5, & I_{\mathrm{cont}} \wedge \operatorname{Acc}(y_0)=0
       \wedge \operatorname{Acc}(y_f)=0,\\
1, & \neg I_{\mathrm{cont}} \wedge \operatorname{Acc}(y_f)=1,\\
0, & \neg I_{\mathrm{cont}} \wedge \operatorname{Acc}(y_f)=0.
\end{cases}
\end{equation}
Full credit is assigned when the inspector correctly finalizes or when continuation corrects an initially wrong answer. If the inspector rejects an incorrect answer but continuation fails to correct it, partial credit separates the inspection decision from the continuation failure. Unnecessary rejection and premature finalization receive no credit. These outcomes score only the completed interaction and never condition the inspection decision in Eq.~\eqref{eq:inspector_information}. For clarity, Eq.~\eqref{eq:correction_reward} presents the \(N_{m}=1\) case; Ablation Study also evaluate \(N_{m}>1\), where the same loop repeats and the reward is computed over the complete trajectory from \(y_0\) to \(y_f\).

The overall reward is:
\begin{equation}\label{eq:role_coupled_reward}
R(\tau)=
\lambda_{\mathrm{acc}}R_{\mathrm{acc}}(\tau)
+\lambda_{\mathrm{fmt}}R_{\mathrm{fmt}}(\tau)
+\lambda_{\mathrm{corr}}R_{\mathrm{corr}}(\tau),
\end{equation}
with $(\lambda_{\mathrm{acc}},\lambda_{\mathrm{fmt}},\lambda_{\mathrm{corr}})=(0.7,0.1,0.2)$, prioritizing correctness while retaining pressure for valid interaction and useful revision. Because $R_{\mathrm{corr}}$ jointly depends on the initial solution, inspection, and continuation outcome, the reward is a coupled function:
\begin{equation}
R(\tau)
=
R\!\left(
\mathcal{M}_{S_1}(\tau),
\mathcal{M}_{I}(\tau),
\mathcal{M}_{S_2}(\tau)
\right),
\end{equation}
rather than a collection of independent role-level objectives.

\subsection{Trajectory-Level Policy Optimization}

The role-coupled reward defines the desired behaviors of all roles, while policy optimization determines how its credit is assigned. For each query $q_b$, the old policy samples $G$ complete rollouts and normalizes their aggregated rewards into group-relative trajectory advantages:
\begin{equation}
\bar A_{b,g}
=
\frac{R(\tau_{b,g})-
\operatorname{mean}_{g'}R(\tau_{b,g'})}
{\operatorname{std}_{g'}R(\tau_{b,g'})+\epsilon}.
\end{equation}
The component rewards are therefore combined before credit assignment rather than routed independently to individual roles. Let $m_{b,g,t}$ mark model-generated tokens, and let $c_{b,g,t}\in\{0,1\}$ specify which of these tokens receive trajectory credit. With $N^c_{b,g}=\sum_t m_{b,g,t}c_{b,g,t}$ and importance ratio $\rho_{b,g,t}=\pi_\theta(a_{b,g,t}\mid h_{b,g,t})/\pi_{\theta_{\mathrm{old}}}(a_{b,g,t}\mid h_{b,g,t})$, the credit-assignment form of GRPO is
Let $\rho_{b,g,t}=\frac{\pi_\theta(a_{b,g,t}\mid h_{b,g,t})}{\pi_{\theta_{\mathrm{old}}}(a_{b,g,t}\mid h_{b,g,t})}$,
$\bar\rho_{b,g,t}=\operatorname{clip}(\rho_{b,g,t},1-\epsilon_c,1+\epsilon_c)$,
and $L_{b,g,t}(\theta)=\min(\rho_{b,g,t}\bar A_{b,g},\bar\rho_{b,g,t}\bar A_{b,g})-\beta D_{\mathrm{KL},b,g,t}$,
\begin{equation}\label{eq:credit_assignment_grpo}
\mathcal{J}(\theta;c)
=
\mathbb{E}_{b,g}\!\left[
\frac{1}{N^c_{b,g}}
\sum_t m_{b,g,t}c_{b,g,t} L_{b,g,t}(\theta)
\right].
\end{equation}
Our full-trajectory assignment sets:
\begin{equation}\label{eq:full_credit_assignment}
{ c^{\mathrm{full}}_{b,g,t}=1,
A^{\mathrm{full}}_{b,g,t}=m_{b,g,t}\bar A_{b,g},
\mathcal{J}_{\mathrm{full}}=\mathcal{J}(\theta;c^{\mathrm{full}}) }.
\end{equation}
Consequently, the aggregated accuracy, validity, and correction signal jointly trains every decision made by the initial solver, inspector, and continuation solver, preserving the complete likelihood in Eq.~\eqref{eq:trajectory_likelihood}.

By contrast, role-selective credit assignment uses a trajectory-dependent mask:
\begin{equation}\label{eq:role_selective_credit}
c^{\mathrm{role}}_{b,g,t}
=z_{b,g,t}(\tau)
=f_t\!\left(y_0,C,y_f,
\operatorname{Acc}(y_0),\operatorname{Acc}(y_f)\right),
\end{equation}
which assigns the advantage only to the role judged locally responsible, such as the inspector when it rejects a correct answer. This is a concrete credit-assignment rule rather than a neutral loss mask. Ignoring clipping and KL regularization, its difference from full-trajectory optimization is:
\begin{equation}\label{eq:masking_bias_main}
g_{\mathrm{role}}-g_{\mathrm{full}}
=
\mathbb{E}_{b,g}\!\left[
\bar A_{b,g}
\left(
\bar s^{\mathrm{role}}_{b,g}
-\bar s^{\mathrm{full}}_{b,g}
\right)
\right],
\end{equation}
where $\bar s^{\mathrm{role}}$ and $\bar s^{\mathrm{full}}$ average token score functions over selected-role and all trainable tokens. Because $z(\tau)$ depends on policy-generated outcomes and the roles causally affect one another, this difference is generally nonzero. Role-selective assignment therefore changes the optimization target and introduces a post-hoc attribution mismatch relative to the full-trajectory GRPO objective, whereas full-trajectory assignment preserves all decisions contributing to the coupled reward and enables joint role adaptation. Supplementary Material~C provides a detailed theoretical derivation and comprehensive experimental comparisons, which collectively demonstrate the effectiveness of our method.

\begin{table}[htbp]
\centering

\setlength{\tabcolsep}{5pt}
\begin{tabular}{lccc}
\toprule
Model & V* Bench & 4K & 8K \\
\midrule
\multicolumn{4}{c}{\textit{\textbf{Direct Answer}}} \\
\midrule
Gemini-2.5-Pro & 83.8 & 87.3 & 85.4 \\
GPT-4o & 67.5 & 65.0 & 59.6 \\
LLaVA-onevision & 75.4 & 63.0 & 59.8 \\
Qwen3-VL-8B-Instruct & 86.4 & 78.9 & 74.6 \\
\midrule
\multicolumn{4}{c}{\textit{\textbf{Agentic Model}}} \\
\midrule
SEAL & 74.8 & - & - \\
Monet & 83.3 & 71.0 & 68.0 \\
Pixel-Reasoner & 84.3 & 72.6 & 66.1 \\
DeepEyes & 83.3 & 73.2 & 69.5 \\
Thyme & 82.2 & 77.0 & 72.0 \\
DeepEyesV2 & 81.8 & 77.9 & 73.8 \\
Mini o3 & 88.2 & 77.5 & 73.3 \\
SenseNova-MARS-8B & 92.2 & \textbf{83.1} & 78.4 \\
\rowcolor{gray!15}
\textbf{MRC-MMAgent-8B} & \textbf{93.2} & 82.1 & \textbf{78.5} \\
\rowcolor{gray!15}
$\Delta$ (vs Qwen3-VL-8B-Instruct) & +6.8 & +3.2 & +3.9 \\
\bottomrule
\end{tabular}
\caption{Accuracy on image reasoning benchmarks. 4K and 8K denote HR-Bench 4K and HR-Bench 8K.}
\label{tab:comparison_hr_bench}
\end{table}

\section{Experiments}

\subsection{Experimental Setup}

\textbf{Benchmarks.}
We evaluate on current knowledge-intensive and visually complex tasks. Knowledge-intensive multimodal search and visual question answering, including FVQA-test~\cite{wu2025mmsearch}, Infoseek~\cite{chen2023can}, SimpleVQA~\cite{cheng2025simplevqa}, MMsearch ~\cite{jiang2024mmsearch}, and LiveVQA~\cite{fu2025livevqa}. Visual grounding and high-resolution perception with V* Bench~\cite{wu2024v} and HR-Bench 4K/8K~\cite{wang2025divide}, testing whether trajectory verification and continuation remain effective beyond search-heavy QA. 



\textbf{Training and Implementation.}
We experiment with Qwen2.5-VL-7B-Instruct~\cite{bai2025qwen2} and Qwen3-VL-8B-Instruct~\cite{yang2025qwen3} multimodal tool agents. Experiments are conducted on $9\times$ NVIDIA H200 GPUs. For fair comparison, we use Qwen3-32B~\citep{yang2025qwen3} as reference DeepEyes2 and SenseNova-MARS to evaluate the answers. We adopt the  three tools: reverse image search, text search, and image zoom. We use a batch size of 128, LR 5e-6, 16K token length, rollout group 8, and 10-turn tool interactions per rollout. The training and test sets are strictly separated with no overlap. The detailed parameter configurations, tool specifications and prompts are listed in Supplementary Material G.

\textbf{Baselines.}
We conduct extensive comparisons against a broad range of baselines, including open-source models (e.g., Qwen~\cite{bai2025qwen2,yang2025qwen3}), closed-source models (e.g., GPT-5~\cite{gpt5}), and open-source multimodal agents (e.g., MMSearch-R1~\citep{wu2025mmsearch}), among others~\cite{gemini3,ops4,yao2026mm,chng2025sensenova,chen2026opensearch,wang2026monet,zhang2025thyme,zheng2025deepeyes}.

\subsection{Main Results}

Table~\ref{tab:main_results} and Figure~\ref{fig:radar_comparison} reports the main multimodal search-intensive results. We compare our method with trained multimodal tool agents at the 7B and 8B scales. Our 7B agent surpasses the trained 7B multimodal frameworks on FVQA-Test, InfoSeek, SimpleVQA, and LiveVQA. At the 8B scale, our method remains highly competitive: compared to SenseNova-MARS-8B, we improve FVQA-Test from $67.1\%$ to $74.2\%$, SimpleVQA from $70.2\%$ to $74.3\%$, and LiveVQA from $56.2\%$ to $61.3\%$. These results demonstrate that explicit trajectory verification and continuation offer complementary advantages for RL-based tool-use training. Notably, the absence of extra datasets for reinforcement training underscores the robustness of our framework. In addition, our method attains the best FVQA-Test and LiveVQA scores among the listed systems, while maintaining competitiveness on others.

Table~\ref{tab:comparison_hr_bench} evaluates whether the proposed workflow also transfers to visual grounding and high-resolution perception. Our method achieves the best V* Bench score and the best HR-Bench 8K score among the listed agentic models. The gains over Qwen3-VL-8B-Instruct are +$6.8\%$ on V* Bench, +$3.2\%$ on HR-Bench 4K, and +$3.9\%$ on HR-Bench 8K. These results suggest that trajectory verification is also effective when the agent must decide whether visual observations are sufficient before finalizing an answer, rather than only in web-style search tasks. Notably,  Supplementary Material H presents a detailed cost comparison, showing that our method achieves leading accuracy with low cost and data efficiency.

\subsection{Ablation Study}

\textbf{Different Workflow Pipeline Performance.}
Table~\ref{tab:dual_agent_comparison} presents a comparison of different reflection framework variants. GRPO, which denotes the single-solver policy under with identical total tokens and tool-call configurations, performs far worse than the multi-role collaborative RL framework. ReflectAgent integrates reflection into the entire chain-of-thought but still performs single-role optimization. However, its performance is comparable to GRPO, and empirical observations indicate that the reflection mechanism tends to degrade during training. The two-stage RL strategy first trains the solution trajectory and answer, followed by an inspection and update of the complete trajectory. Although this approach leads to certain improvements, it incurs substantially higher training time costs. Our framework effectively overcomes these limitations and achieves optimal performance. Furthermore, when the role-coupled reward is removed, leaving only $R_{\mathrm{acc}}$ and the format reward, we observe a clear drop in model performance. This is because removing the role-coupled reward weakens the overall normative constraints of the multi-role framework, thereby hindering effective optimization of multi-role coordination. resulting in a performance decline on FVQA-Test (from $64.0\%$ to $62.3\%$).

\begin{table}[htbp]
\centering

\setlength{\tabcolsep}{3pt}
\begin{tabular}{lcc}
\toprule
Method & FVQA-Test & MMSearch \\
\midrule
GRPO & 58.3 & 50.9 \\
ReflectAgent & 58.6 & 52.3 \\
Solver-Inspector(two-stage RL) & 59.1 & 56.0 \\
Ours w/o role-coupled reward & 62.3 & 60.2 \\
\rowcolor{gray!10}
Ours & \textbf{64.0} & \textbf{62.0} \\
\bottomrule
\end{tabular}
\caption{The ablation of different inspect and solve workflow with 7B.}
\label{tab:dual_agent_comparison}
\end{table}

\begin{figure}[htb]
\centering
\includegraphics[width=0.45\textwidth]{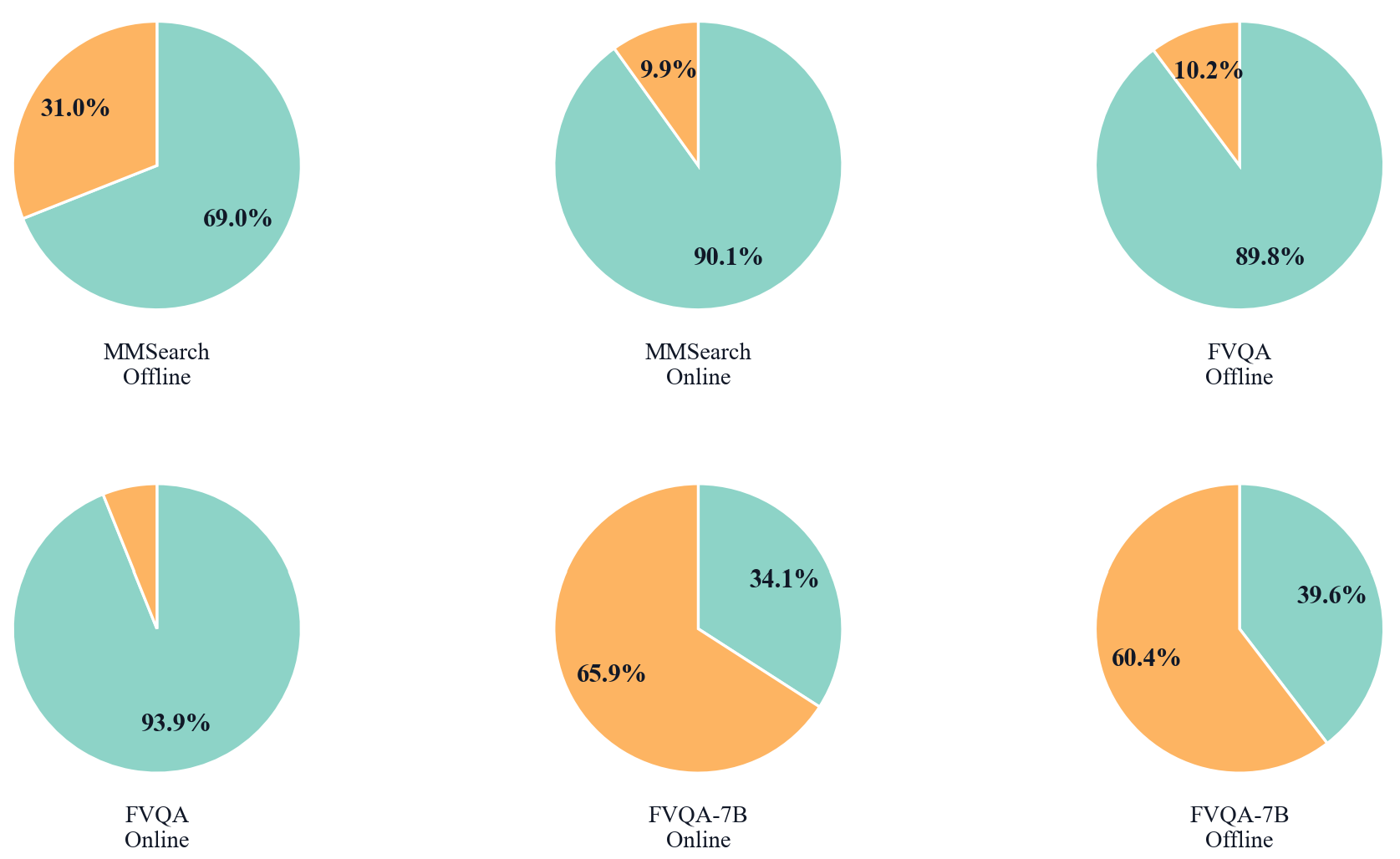}
\caption{Pie chart visualizations comparing ablation results across model sizes, training approaches, and benchmarks. Orange indicates REJECT, and green indicates PASS.}
\label{fig:inspector_decision_pies}
\end{figure}
\textbf{Experiments on Framework Adaptability.}
Table~\ref{tab:search_configuration_behavior} and Figure~\ref{fig:inspector_decision_pies} present ablation studies under various data configurations, training settings, and model scales. The inspection capability adapts across model sizes; the 8B model achieves a higher one-pass success rate, consistent with stronger models facing easier tasks. Online search outperforms offline search, leading to improved one-pass accuracy. The framework generalizes well across benchmarks of varying difficulty, and on V*Bench, where most trajectories are finalized after the initial solve, full-trajectory credit assignment still reinforces the solver's decisions, improving one-pass reasoning without relying on continuation-based correction.

\begin{table}[htbp]
\centering

\setlength{\tabcolsep}{7pt}
\begin{tabular}{lccc}
\toprule
Dataset & Training & Finalize & Continue \\
\midrule
MMSearch & Offline & 69.0 & 31.0 \\
MMSearch & Online & \textbf{90.1} & \textbf{9.9} \\
FVQA & Offline & 89.8 & 10.2 \\
FVQA & Online & \textbf{93.9} & \textbf{6.1} \\
FVQA-7B & Offline & 34.1 & 65.9 \\
FVQA-7B & Online & 39.6 & 60.4 \\
V* Bench& -- & 99.8 & 0.2 \\
\bottomrule
\end{tabular}
\caption{Ablation studies on model scales, training methods, and benchmarks. Finalize indicates that the inspection is passed and the task is completed. Continue indicates that issues exist in the current trajectory and further problem-solving is required. Offline uses the local knowledge base Wiki25, while online search uses the more comprehensive Google Search.}
\label{tab:search_configuration_behavior}
\end{table}

\begin{table}[htbp]
\centering

\setlength{\tabcolsep}{2pt}
\begin{tabular}{lcccccc}
\toprule
Setting & Solver & Inspector & Solver1 & Solver2 & Err. & Acc \\
\midrule
\rowcolor{gray!10}
Config. 1 & $\checkmark$ & $\checkmark$ & \textbf{57.2} & \textbf{72.6} & \textbf{5.3} & \textbf{64.0} \\
Config. 2 & $\checkmark$ & $\times$ & 54.3 & 72.3 & 10.4 & 62.2 \\
Config. 3 & $\times$ & $\checkmark$ & 41.1 & 53.7 & 14.7 & 52.9 \\
Config. 4 & $\times$ & $\times$ & 18.4 & 52.9 & 22.6 & 51.1 \\
\bottomrule
\end{tabular}
\caption{Functional ablation under different role-training settings with 7B. $\checkmark$ indicates that the corresponding role is trained, and $\times$ indicates that it is frozen.}
\label{tab:function_ablation}
\end{table}

\textbf{The Ablation Implementation of Multi-Role Training Configuration.}
To further analyze the impact of individual role training on the overall performance, and to validate the effectiveness of our method, we conduct component-level ablation studies on the training states of the Solver and Inspector. In Table~\ref{tab:function_ablation}, Config. 1 trains both solving and inspection roles, Config. 2 trains only the solving roles, Config. 3 trains only the inspection role, and Config. 4 trains neither component. Our framework consists of two solving stages, denoted as Solver1 and Solver2, and one inspection stage. The Solver1 metric measures the probability that the first solving stage answers correctly. The Solver2 metric measures the probability that the agent produces the correct answer after second-stage solving or revision. The inspection error metric measures the proportion of incorrect control decisions, where a lower value indicates more reliable trajectory assessment. Overall accuracy reports the final task accuracy of the complete pipeline.

From an overall perspective, even when neither the solving nor inspection role is trained, the framework still performs better than direct inference, suggesting that the solve--inspect--continue process itself is beneficial. As more functional components are trained, performance improves steadily: overall accuracy increases from $51.1\%$ to $52.9\%$ and $62.2\%$, and reaches $64.0\%$ when solving and inspection are jointly trained. This trend shows that the gain does not come from a single isolated module, but from the coordinated improvement of initial solving, trajectory assessment, and second-stage correction. Config. 4 also has weaker format compliance, so we report accuracy on valid trajectories. The interaction between solving and inspection further supports this conclusion. When only the solving roles are trained, Solver2 reaches $72.3\%$, indicating strong second-stage correction, but inspection error remains $10.4\%$ due to untrained inspection, limiting final accuracy. When only the inspection role is trained, inspection error drops significantly, but Solver1 and Solver2 remain weaker, showing that reliable assessment must be paired with a solver that can act on feedback. When both are jointly trained, Solver1, Solver2, and inspection error all achieve the best or near-best results. This demonstrates that the role-coupled reward simultaneously improves first-attempt gathering, trajectory assessment, and post-inspection correction, leading to the best overall task accuracy.

\textbf{Robustness and Effectiveness Analysis.}
We conduct extensive ablation studies, with all results reported over the first 100 training steps to mitigate the influence of convergence fluctuations. Supplementary Material D investigates the effect of the number of inspection-refinement loops. The accuracy curves for the single-loop (\(N_m=1\)) and two-loop (\(N_m=2\)) settings almost completely overlap, yielding  accuracies of 50.9$\%$ and 50.3$\%$, respectively---indicating highly comparable performance. However, the two-loop setting increases the average output length from 1,717.5 to 2,751.7 tokens and the average number of tool calls from 2.5 to 2.9, incurring substantial computational overhead. This suggests that additional loops introduce contextual redundancy rather than performance gains; therefore, we adopt \(N_m=1\) as the default configuration to maintain efficiency without sacrificing performance. Furthermore, compared with the baseline, our method achieves notable accuracy improvements with only a marginal increase in token and tool-call costs. This indicates that our approach does not introduce appreciable latency while delivering superior performance, and that it effectively helps the model avoid redundant attempts. Supplementary Material E compares the Inspector with and without training. The untrained Inspector marks nearly all trajectories as Continue (98.3$\%$ continue rate), leading to numerous unnecessary revisions and an accuracy of only 62.7$\%$. In contrast, the trained Inspector reduces the continue rate to 60.4$\%$ and improves accuracy to 64.0$\%$, demonstrating that our multi-role training framework endows the Inspector with effective reasoning capability. Supplementary Material F presents ablations on the role-coupled reward \(R_{\mathrm{corr}}\). Over the first 100 training steps, different weight configurations (0.1, 0.2, and 0.3) achieve accuracies of 63.1$\%$, 64.7$\%$, and 63.6$\%$, respectively, all substantially outperforming the baseline of 57.3$\%$ with consistent convergence speed. This confirms that our method exhibits strong robustness across various reward settings and that the performance improvements do not depend on specific hyperparameter choices.

\section{Conclusion}
We propose a multi-role collaborative reinforcement learning framework that internalizes trajectory assessment and continuation decisions into the executable agent loop, enabling multimodal tool agents to achieve leading or highly competitive performance across multiple benchmarks without requiring additional training data.

\bibliography{aaai2027}
\newpage
\newpage
\appendix

\section{A.Structural Comparison with Representative Reflective and Tool-use RL
Methods.}
As summarized in Table~\ref{tab:novelty_positioning}, our method is clearly distinct from current approaches. Specifically, we train multiple multimodal roles simultaneously in a synchronized manner. The Inspector operates in an unsupervised fashion, which aligns closely with real-world application scenarios. Moreover, our approach achieves highly competitive performance without requiring any additional data.
\begingroup
\newcommand{\fullmark}{\ensuremath{\checkmark}}
\newcommand{\partmark}{\ensuremath{\triangle}}
\newcommand{\nomark}{\ensuremath{\times}}

\begin{table*}[t]
\centering
\setlength{\tabcolsep}{4.0pt}
\renewcommand{\arraystretch}{1.18}
\begin{tabular}{@{}lccccc@{}}
\toprule
\textbf{Method}
& \shortstack{\textbf{No extra}\\\textbf{SFT trajectories}}
& \shortstack{\textbf{Trainable}\\\textbf{control inspector}}
& \shortstack{\textbf{Reference-free}\\\textbf{inspection}}
& \shortstack{\textbf{Inspection-triggered}\\\textbf{continuation}}
& \shortstack{\textbf{Full-trajectory}\\\textbf{coupled update}} \\
\midrule
Look-Back
& \nomark & \nomark & \nomark & \nomark & \nomark \\
Klissarov et al.(Google)
& \nomark & \partmark & \nomark & \fullmark & \nomark \\
VSearcher
& \nomark & \nomark & \nomark & \nomark & \nomark \\
PyVision-RL
& \partmark & \nomark & \nomark & \nomark & \nomark \\
Tool-Star
& \nomark & \partmark & \partmark & \nomark & \nomark \\
VideoChat-M1
& \partmark & \nomark & \nomark & \nomark & \partmark \\
ReGRPO
& \nomark & \nomark & \nomark & \nomark & \fullmark \\
\textbf{MRC-MMAgent (ours)}
& \textbf{\fullmark} & \textbf{\fullmark} & \textbf{\fullmark}
& \textbf{\fullmark} & \textbf{\fullmark} \\
\bottomrule
\end{tabular}
\caption[Structural comparison with representative reflective and tool-use RL
methods.]{Structural comparison with representative reflective, multimodal
tool-use, and multi-role RL methods. \fullmark: explicit support;
\partmark: partial or indirect support; \nomark: absent or not a primary
training objective. Reference-free inspection means that the inspection
decision does not use a teacher critique or ground-truth answer. Full-trajectory
coupled update means that one trajectory reward updates all policy-generated
role tokens without post-hoc role-selective masking.}
\label{tab:novelty_positioning}
\end{table*}

\endgroup
\section{B. Pseudo algorithm for Multi Role Tool Agent Rollout}
Algorithm 1 illustrates how multiple roles collaborate to ensure the successful completion of multimodal agent tasks.

\section{C. Why Full-Trajectory GRPO Is Preferable to Role-Selective Masking}

This explains why full-trajectory GRPO is better aligned with our multi-role agent loop than role-selective masking. The derivation follows the likelihood-ratio policy-gradient view of REINFORCE, the clipped surrogate of PPO, and the group-relative advantage construction of GRPO. The core issue is that the solver, inspector, and continuation solver are not independent modules with separable rewards. They are prompt-conditioned behaviors of the same policy, and the final reward is a coupled function of all role decisions.

\subsection{Full-Trajectory Objective}

For a query $q_b$, GRPO samples a group of $G$ trajectories from the old policy:
\begin{equation}
\{\tau_{b,g}\}_{g=1}^{G}
\sim
\pi_{\theta_{\mathrm{old}}}(\cdot\mid q_b).
\end{equation}
The trainable model tokens in a trajectory are
\begin{equation}
\mathcal{M}(\tau)
=
\mathcal{M}_{S_1}(\tau)
\cup
\mathcal{M}_{I}(\tau)
\cup
\mathcal{M}_{S_2}(\tau),
\end{equation}
where $\mathcal{M}_{S_1}$, $\mathcal{M}_{I}$, and $\mathcal{M}_{S_2}$ denote the first solver, inspector, and continuation solver tokens. System prompts, tool observations, and bridge messages are conditioning context and receive no likelihood loss.

The role-coupled trajectory reward is
\begin{equation}
R(\tau)
=
\lambda_{\mathrm{acc}}R_{\mathrm{acc}}(\tau)
+
\lambda_{\mathrm{fmt}}R_{\mathrm{fmt}}(\tau)
+
\lambda_{\mathrm{corr}}R_{\mathrm{corr}}(\tau),
\end{equation}
where $R_{\mathrm{acc}}$ evaluates final answer correctness, $R_{\mathrm{fmt}}$ evaluates interaction validity, and $R_{\mathrm{corr}}$ evaluates whether inspection-triggered continuation is useful.

Following GRPO, we compute a group-relative advantage:
\begin{equation}
A_{b,g}
=
\frac{
R(\tau_{b,g})
-
\frac{1}{G}\sum_{g'=1}^{G}R(\tau_{b,g'})
}{
\operatorname{Std}_{g'}[R(\tau_{b,g'})]+\epsilon
}.
\end{equation}
Under outcome supervision, GRPO assigns the normalized outcome reward to all sampled output tokens. Analogously, we assign the scalar trajectory advantage to all trainable model tokens:
\begin{equation}
A_{b,g,t}=A_{b,g}m_{b,g,t},
\end{equation}
where $m_{b,g,t}=1$ for trainable model-generated tokens and $m_{b,g,t}=0$ for non-trainable context or padding tokens.

Ignoring clipping and KL regularization for clarity, the full-trajectory update has the form
\begin{equation}
g_{\mathrm{full}}
=
\mathbb{E}_{b,g}
\left[
\frac{A_{b,g}}{N_{b,g}}
\sum_t
m_{b,g,t}
\nabla_\theta
\log\pi_\theta(a_{b,g,t}\mid h_{b,g,t})
\right],
\end{equation}
where
\begin{equation}
N_{b,g}=\sum_t m_{b,g,t}.
\end{equation}
With PPO-style clipping and KL regularization, the corresponding GRPO surrogate can be written as
\begin{equation}
\mathcal{J}_{\mathrm{full}}(\theta)
=
\mathbb{E}_{b,g}
\left[
\frac{1}{N_{b,g}}
\sum_t
m_{b,g,t}
\left(
\ell^{\mathrm{clip}}_{b,g,t}(\theta)
-
\beta D_{\mathrm{KL},b,g,t}
\right)
\right].
\label{eq:full_grpo}
\end{equation}
where $\ell^{\mathrm{clip}}_{b,g,t}(\theta)$ denotes the
PPO-style clipped surrogate objective:
\begin{equation}
\ell^{\mathrm{clip}}_{b,g,t}(\theta)
=
\min
\left(
\rho_{b,g,t}A_{b,g},
\bar{\rho}_{b,g,t}A_{b,g}
\right),
\label{eq:clipped_surrogate}
\end{equation}
where the clipped importance ratio is defined as
\begin{equation}
\bar{\rho}_{b,g,t}
=
\operatorname{clip}
\left(
\rho_{b,g,t},
1-\epsilon_c,
1+\epsilon_c
\right).
\label{eq:clipped_ratio}
\end{equation}

This matches the reward structure: because $R(\tau)$ scores the complete solve--inspect--continue process, all model decisions contributing to the outcome receive the same group-relative signal.

\subsection{Role-Selective Masking Objective}

A role-selective masking variant introduces an additional binary mask $z_{b,g,t}$ based on post-hoc role-level correctness, e.g., updating only inspector tokens when the first solver is correct but the inspector rejects, or updating only solver tokens when the inspector is locally correct but the final correction fails.

This changes the GRPO surrogate to
\begin{equation}
\mathcal{J}_{\mathrm{mask}}(\theta)
=
\mathbb{E}_{b,g}
\left[
\frac{1}{N^z_{b,g}}
\sum_t
m_{b,g,t}z_{b,g,t}
\widetilde{\ell}_{b,g,t}(\theta)
\right].
\label{eq:masked_grpo}
\end{equation}
where $\widetilde{\ell}_{b,g,t}(\theta)$ denotes the
KL-regularized clipped surrogate:
\begin{equation}
\widetilde{\ell}_{b,g,t}(\theta)
=
\ell^{\mathrm{clip}}_{b,g,t}(\theta)
-
\beta D_{\mathrm{KL},b,g,t}.
\label{eq:regularized_surrogate}
\end{equation}
where
\begin{equation}
N^z_{b,g}=\sum_t m_{b,g,t}z_{b,g,t}.
\end{equation}
This matches our mask implementation: the role mask directly modifies the token-level loss mask, i.e., the effective loss mask is $m_{b,g,t}z_{b,g,t}$.
The corresponding unclipped gradient form is
\begin{equation}
g_{\mathrm{mask}}
=
\mathbb{E}_{b,g}
\left[
\frac{A_{b,g}}{N^z_{b,g}}
\sum_t
m_{b,g,t}z_{b,g,t}
s_{b,g,t}(\theta)
\right].
\label{eq:masked_gradient}
\end{equation}
where $s_{b,g,t}(\theta)$ denotes the token-level score function:
\begin{equation}
s_{b,g,t}(\theta)
=
\nabla_\theta
\log
\pi_\theta
\left(
a_{b,g,t}\mid h_{b,g,t}
\right).
\label{eq:token_score}
\end{equation}
\subsection{Why Masking Introduces Role-Attribution Bias}

The difference between the masked and full updates is
\begin{equation}
g_{\mathrm{mask}}-g_{\mathrm{full}}
=
\mathbb{E}_{b,g}
\left[
A_{b,g}
\left(
\bar{s}^{\mathrm{mask}}_{b,g}
-
\bar{s}^{\mathrm{full}}_{b,g}
\right)
\right].
\label{eq:gradient_difference}
\end{equation}
where $\bar{s}^{\mathrm{mask}}_{b,g}$ is the masked average
score function:
\begin{equation}
\bar{s}^{\mathrm{mask}}_{b,g}
=
\frac{1}{N^z_{b,g}}
\sum_t
m_{b,g,t}z_{b,g,t}
s_{b,g,t}(\theta),
\label{eq:masked_average_score}
\end{equation}
and $\bar{s}^{\mathrm{full}}_{b,g}$ is the corresponding
full-trajectory average:
\begin{equation}
\bar{s}^{\mathrm{full}}_{b,g}
=
\frac{1}{N_{b,g}}
\sum_t
m_{b,g,t}
s_{b,g,t}(\theta).
\label{eq:full_average_score}
\end{equation}
In general, this term is nonzero. Thus, masking is not a lower-variance version of the full objective; it changes the optimization target.

To see this more directly, ignore the normalization difference between $N^z_{b,g}$ and $N_{b,g}$. Then the masking bias is approximately
\begin{equation}
\operatorname{Bias}_{\mathrm{mask}}
\approx
-
\mathbb{E}_{b,g}
\left[
A_{b,g}
\sum_t
m_{b,g,t}(1-z_{b,g,t})
s_{b,g,t}(\theta)
\right].
\label{eq:mask_bias}
\end{equation}

This bias vanishes only if the removed tokens are uncorrelated with the trajectory advantage:
\begin{equation}
\mathbb{E}_{b,g}
\left[
A_{b,g}
\sum_{t:z_{b,g,t}=0}
\nabla_\theta\log\pi_\theta(a_{b,g,t}\mid h_{b,g,t})
\right]
=0.
\end{equation}
This condition does not hold in our setting: first-solver evidence affects inspection, inspector feedback affects continuation, and continuation success determines whether rejection was useful. Hence masked tokens can still have causal and statistical influence on the final reward.

\subsection{Masking Is Not a Baseline}

The unbiasedness of a standard policy-gradient baseline relies on the fact that the baseline does not depend on the sampled action:
\begin{equation}
\mathbb{E}_{\tau}
\left[
b(q)\nabla_\theta\log p_\theta(\tau\mid q)
\right]
=
b(q)\nabla_\theta
\sum_{\tau}p_\theta(\tau\mid q)
=0.
\end{equation}
Thus subtracting an action-independent baseline can reduce variance without changing the expected gradient.

Role-selective masking is different. The mask is a post-hoc function of the sampled trajectory:
\begin{equation}
z_{b,g,t}
=
f(y_0,\ C,\ y_f,\ A(y_0),\ A(y_f)),
\end{equation}
where $y_0$ is the initial answer, $C$ is the inspector decision, and $y_f$ is the final answer. These quantities are generated by the policy itself, so $z_{b,g,t}$ is action-dependent. It cannot be treated as variance-reducing baseline; it removes likelihood-gradient terms and changes the expected update.

This is consistent with standard baseline theory: action-independent or state-dependent baselines can reduce variance without changing the expected policy gradient, while action-dependent control variates require a correction term to remain unbiased Role-selective masking has no such correction.

\subsection{Failure Modes of Role-Selective Masking}

\paragraph{Correct initial answers may still need solver-side credit.}
If the first solver answers correctly but the inspector incorrectly rejects, masking may update only the inspector. Yet the solver may have produced weak or ambiguous evidence, making the answer hard to verify. Full-trajectory GRPO can reinforce both better inspection and more verifiable solver trajectories.

\paragraph{Correct rejection does not imply useful feedback.}
If the first solver is wrong, the inspector correctly rejects, but continuation fails, masking the inspector assumes the inspector is not responsible. However, the rejection reason may be uninformative. Full-trajectory optimization encourages feedback that helps downstream correction, not merely a locally correct label.

\paragraph{Relative advantages can reinforce the wrong local component.}
GRPO advantages are group-relative. A trajectory with an imperfect local decision can still receive positive advantage if it outperforms other samples. Masking may preserve only selected role tokens and assign them this advantage, even when the selected local action was not the source of improvement.

\paragraph{Tool-use credit is weakened.}
Tool calls are primarily solver tokens. If useful tool use is followed by an inspector error, masking may remove solver tokens and erase positive tool-use credit. Failed trajectories may still keep solver tokens with negative advantage, reducing tool invocation frequency.

\paragraph{Conclusion}

The full trajectory likelihood decomposes as
\begin{equation}
\log \pi_\theta(\tau)
=
\sum_{X\in\{S_1,I,S_2\}}
\sum_{t\in\mathcal{M}_X}
\log \pi_\theta(a_t\mid h_t).
\label{eq:trajectory_logprob}
\end{equation}
Since the reward is also a coupled function of these roles,
\begin{equation}
R(\tau)
=
R(\mathcal{M}_{S_1},\mathcal{M}_{I},\mathcal{M}_{S_2}),
\end{equation}
the optimization objective should preserve the full likelihood term. Full-trajectory GRPO assigns the group-relative trajectory advantage to all model-generated tokens. Role-selective masking instead optimizes
\begin{equation}
\log\pi_\theta^{\mathrm{mask}}(\tau)
=
\sum_{t\in\mathcal{M}(\tau)}
z_t(\tau)\log\pi_\theta(a_t\mid h_t),
\end{equation}
which introduces post-hoc role-attribution bias and breaks the alignment between the coupled reward and the optimized likelihood. This explains why full-trajectory joint optimization is more effective in our multi-role agent framework.

\subsection{The Effectiveness of  Objective Optimization.}
\begin{figure}[t]
    \centering
    \begin{subfigure}[b]{0.23\textwidth}  
        \centering
        \includegraphics[width=\linewidth]{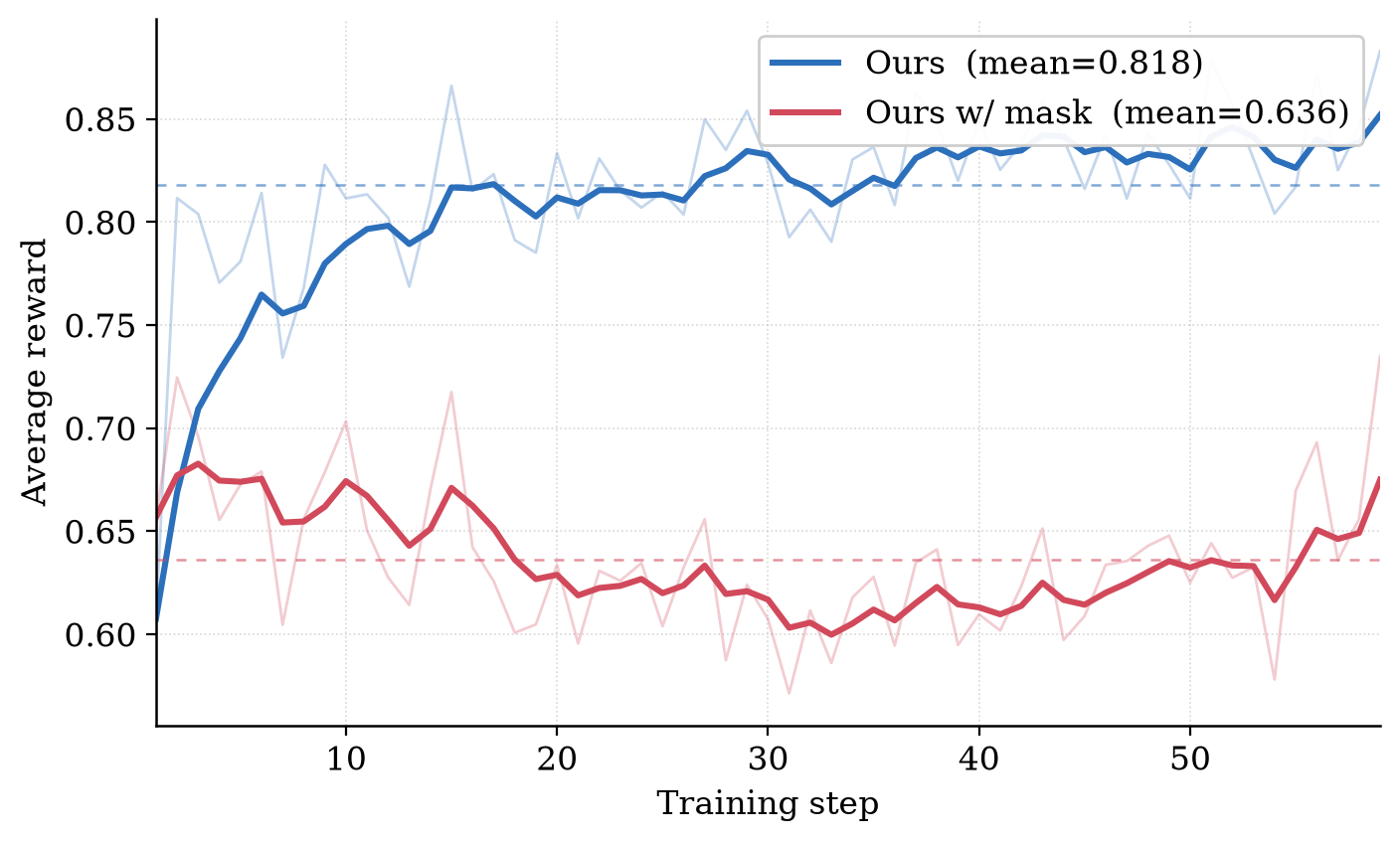}
    \end{subfigure}
    \hfill
    \begin{subfigure}[b]{0.23\textwidth}  
        \centering
        \includegraphics[width=\linewidth]{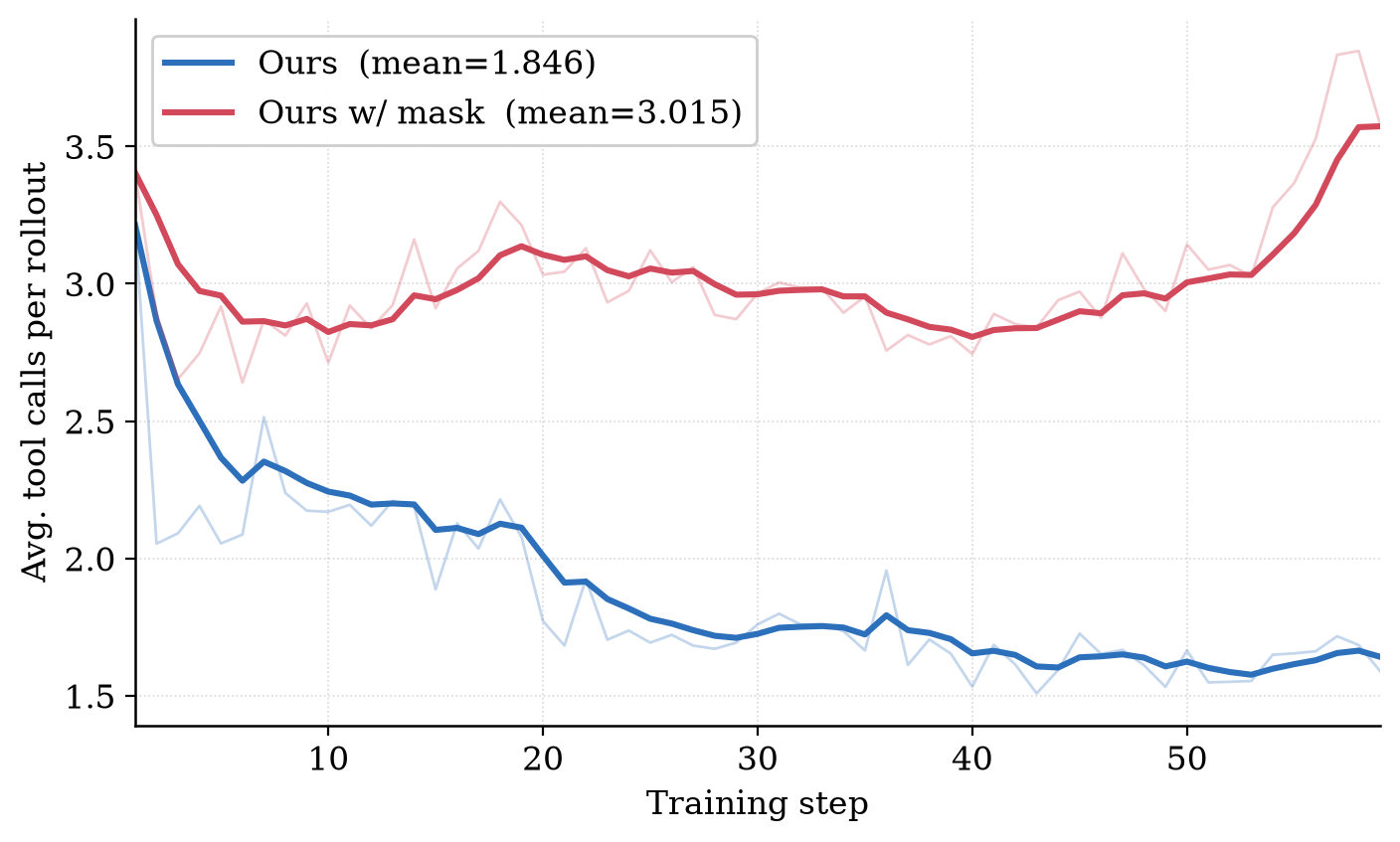}
    \end{subfigure}
    \caption{Comparison of rewards and tool calls for our method and mask-based variant for role-selective optimization.}
    \label{fig:5}
\end{figure}

To support the effectiveness of our objective optimization method, we conduct a comparative analysis with an ideal mask-based optimization training variant. In this variant, we apply selective gradient updates based on the correctness of the Solver and the Inspector: when the Solver makes an error and the Inspector correctly detects it, only the solution trajectory is optimized; when the Solver is correct but the Inspector misjudges it, only the inspection capability is optimized, and so forth. As illustrated in Figure~\ref{fig:5}, the mask-based optimization leads to a substantial drop in training rewards and a reduction in tool invocation counts. These results suggest that our full-trajectory joint training is more conducive to optimizing model performance from a global perspective, further demonstrating the high compatibility between our overall framework and the joint reward design.

\subsection{D. Ablation on the Number of Continuation Loops}

To study whether multiple inspection-refinement loops are necessary, we compare our default single-continuation setting with a two-continuation variant. In our framework, the model first produces an initial answer, then the inspector
checks whether the current response contains remaining issues. If the response passes the inspection, the process terminates; otherwise, the model continues solving the problem and produces a revised answer. This forms one continuation
loop. The two-continuation variant repeats this inspection-refinement process once more.

As shown in Figure~\ref{tab:continuation_ablation} and Table~\ref{tab:continuation_ablation}, the single-continuation and multi-continuation settings achieve broadly comparable performance, with two turns performing slightly worse while incurring substantially longer context and more tool calls. Specifically, the average output length increases from 1,717.50 to 2,751.66 tokens, and the average number of tool calls rises from 2.53 to 2.90. These results suggest that the additional context from more turns may introduce redundancy and potentially increase hallucination. A single inspection turn is sufficient to achieve optimal performance while significantly reducing computational cost, further demonstrating the effectiveness and robustness of our method.
\begin{algorithm}[t]
\caption{Multi-Role Tool-Agent Rollout}
\label{alg:rollout}
\begin{algorithmic}[1]
\STATE Init $h$ with query $q=(x,u)$, tool schemas, solver prompt $S_1$.
\STATE $j\leftarrow 1,\; n_{\mathrm{cont}}\leftarrow 0$.
\WHILE{budget remains}
    \STATE Sample action $a\sim\pi_\theta(\cdot\mid h)$.
    \IF{action is tool call}
        \STATE Execute tool, append observation to $h$.
    \ELSIF{action is candidate answer}
        \STATE Append inspector prompt $S_3$ to $h$.
        \STATE Sample $c\sim\pi_\theta(\cdot\mid h)$.
        \IF{$c$ finalizes}
            \STATE terminate.
        \ELSIF{$c$ requests cont. and $n_{\mathrm{cont}}<=N_m$}
            \STATE $n_{\mathrm{cont}}\leftarrow n_{\mathrm{cont}}+1,\; j\leftarrow j+1$.
            \STATE Append bridge feedback and $S_2$ to $h$.
        \ELSE
            \STATE terminate.
        \ENDIF
    \ENDIF
\ENDWHILE
\end{algorithmic}
\end{algorithm}

\begin{table}[t]
\centering

\begin{tabular}{lccc}
\toprule
$N_m$ & Accuracy & Avg. Tokens & Avg. Tool Calls \\
\midrule
1 & \textbf{50.9} & \textbf{1717.5} & \textbf{2.5} \\
2 & 50.3 & 2751.7 & 2.9 \\
Baseline & 45.6 & 1643.4 & 2.3 \\
\bottomrule
\end{tabular}
\caption{Ablation study on the number of continuation loops with 8B in offline.}
\label{tab:continuation_ablation}
\end{table}

\begin{figure}[htb]
\centering
\includegraphics[width=0.45\textwidth]{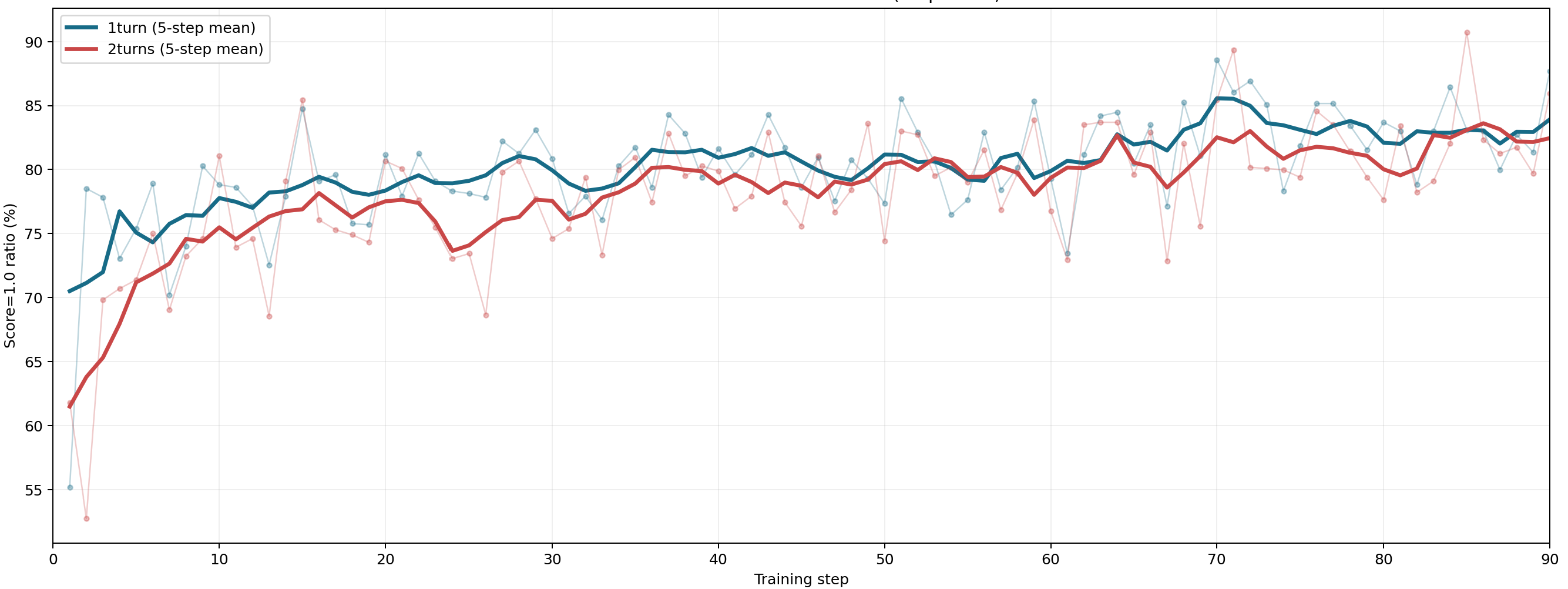}
\caption{Mean Acc on the number of continuation loops.}
\label{fig:turns_abation}
\end{figure}

We attribute this phenomenon to the non-monotonic nature of model-based inspection and refinement. Since the inspector is also implemented by the model, it may occasionally reject an already correct or sufficiently reliable answer.
Additional continuation loops can amplify such false rejections, causing the model to perform unnecessary search, reasoning, or rewriting. This may introduce answer drift and accumulate tool-use noise. Moreover, later refinement states are
generated from already-modified intermediate responses, which are more likely to deviate from the training distribution and thus become less stable.

This ablation demonstrates an important advantage of our design: the proposed framework does not rely on repeatedly applying expensive self-correction loops. Instead, a single targeted inspection-refinement step is sufficient to capture
most correctable errors while avoiding over-correction. Therefore, our single-continuation design provides a better accuracy-cost trade-off and improves inference efficiency.

\subsection{E. Trained vs. Untrained Inspection.}
Table~\ref{tab:checker_behavior} compares the behavioral capabilities of the Inspector. The untrained variant without reasoning capabilities almost consistently marks every trajectory as fail, which introduces numerous unnecessary modifications and consequently degrades performance. In contrast, the trained framework achieves the best accuracy while maintaining a more selective continuation profile. This demonstrates that through our multi-role training framework, the Inspector acquires robust reasoning capabilities, resulting in significantly improved performance.
\begin{table}[htbp]
\centering

\setlength{\tabcolsep}{5pt}
\begin{tabular}{lccc}
\toprule
Method & Acc & Finalize & Continue \\
\midrule
Config. 2 w/ think & 62.2 & 31.5 & 68.5 \\
Config. 2 w/o think & 62.7 & 1.7 & 98.3 \\
\rowcolor{gray!10}
Ours & \textbf{64.0} & \textbf{39.6} & 60.4 \\
\bottomrule
\end{tabular}
\caption{The abation of inspector's action on FVQA-Test.}
\label{tab:checker_behavior}
\end{table}

\subsection{F. Ablation Study on  Role-coupled Reward.}
Table~\ref{tab:reward_aba}  presents the ablation study on our multi-role reward. It can be observed that regardless of the reward ablation, the model consistently and substantially outperforms the baseline, demonstrating the strong robustness of our method.
\begin{table}[htbp]
\centering

\begin{tabular}{lccc}
\toprule
Setting & Acc & Step  \\
\midrule
0.2 & 64.7 & 100  \\
0.3 & 63.6 & 100  \\
0.1 & 63.1 & 100 \\
Baseline & 57.3 & 100 \\
\bottomrule
\end{tabular}
\caption{The abation of inspector's action on FVQA-Test with 8B in offline.}
\label{tab:reward_aba}
\end{table}

\begingroup
\newcommand{\fullmark}{\ensuremath{\checkmark}}
\newcommand{\partmark}{\ensuremath{\triangle}}
\newcommand{\nomark}{\ensuremath{\times}}

\begin{table*}[t]
\centering
\small
\setlength{\tabcolsep}{4pt}
\renewcommand{\arraystretch}{1.18}
\begin{tabular}{@{}lcccccc@{}}
\toprule
\textbf{Method}
& \textbf{SFT}
& \textbf{New Training Dataset}
& \textbf{Base Dataset}
& \textbf{Train-time Web Search}
& \textbf{Test-time Web Search}
& \textbf{ACC} \\
\midrule
MMSearch-R1
& \nomark & \nomark & \fullmark & \fullmark & \fullmark & 51.1 \\

DeepEyes2-7B
& \fullmark & \fullmark & \fullmark & \fullmark & \fullmark & -- \\

MM-DeepResearch-7B
& \fullmark & \fullmark & \fullmark& \nomark & \fullmark & 60.8 \\
\textbf{MRC-MMAgent-7B (ours)}
& \textbf{\nomark}
& \textbf{\nomark}
& \textbf{\fullmark}
& \textbf{\nomark}
& \textbf{\fullmark}
& \textbf{62.2} \\
SenseNova-MARS-8B
& \fullmark & \fullmark & \fullmark& \nomark & \fullmark & 64.6 \\

OpenSearch-VL-8B
& \fullmark & \fullmark & \fullmark& \nomark & \fullmark & 67.5 \\



\textbf{MRC-MMAgent-8B (ours)}
& \textbf{\nomark}
& \textbf{\nomark}
& \textbf{\fullmark}
& \textbf{\nomark}
& \textbf{\fullmark}
& \textbf{68.1} \\
\bottomrule
\end{tabular}

\caption{Comparison of training data, web-search usage, and accuracy among
representative methods. \fullmark: used; \partmark: partially used;
\nomark: not used. ACC denotes the average accuracy across the evaluated
benchmarks.}
\label{tab:chengbenfenxi}
\end{table*}

\endgroup
\subsection{G. Detailed experimental parameter settings.}
 After experimental verification, we set $N_m$ to 1. Following our training script, we use a training batch size of 128, validation batch size of 512, learning rate $5\times10^{-6}$, PPO mini-batch size 128, and micro-batch size 2 per GPU. The maximum prompt and response lengths are both 16K tokens, the rollout group size is 8, and each multi-turn rollout allows up to 10 assistant turns with a maximum tool-response length of 8192 tokens. We use bfloat16 FSDP training, gradient checkpointing, tensor parallel size 2 for rollout, Ulysses sequence parallel size 2, and asynchronous SGLang rollout. During validation, we report task accuracy using the answer evaluator without process reward terms. 
For image search, we cache results from Google Search, each returning five images with their corresponding titles. For text search during training, we build a local knowledge base by chunking and indexing the Wiki 2025 corpus; during testing, we use the online Google Search API, which incurs low usage cost. For training, we only use the original training sets from MMSearch-R1 and DeepEyes2, without introducing any additional training data, and ensure that training and test sets are strictly separated. All ablation studies and baseline comparisons are conducted under identical configurations. Specific prompts can refer to Supplementary Material I.

\subsection{H. Cost-Performance Analysis}
Taking the average accuracy on search-intensive tasks as an example, Table~\ref{tab:chengbenfenxi} presents a cost-performance comparison between our method and current multimodal agent approaches. It can be observed that our method achieves highly competitive performance at significantly lower cost. For instance, our method incurs lower cost than MMSearch-R1 while outperforming it by 11.1$\%$ in accuracy.

\subsection{I. Prompt Templates for the Multi-Role Agent Loop}

\paragraph{Solver 1 Prompt.}
This prompt is used as the initial solver system prompt.
\begin{lstlisting}[
    basicstyle=\footnotesize\ttfamily,
    breaklines=true,
    breakatwhitespace=false,
    columns=fullflexible,
    keepspaces=true,
    numbers=none,
    frame=single,
    caption={Solver 1 system prompt.}
]
#Role
You are a step-by-step reasoning assistant. Your task is to solve the problem **one substep at a time** by analyzing provided information and using available tools. 

## Guiding Principles
1. **Thought-First:** All outputs MUST begin with a thought enclosed in <thinking> </thinking> tags. Explain your current reasoning, what information is missing, and what to do next. Do not output the previous reasoning chain.
2. **One Action per Turn:** At each turn, you must EITHER: 
   - Issue **one specific tool call** enclosed in <tool_call> </tool_call> tags, 
   - OR provide the **final answer** enclosed in <answer> </answer> tags.
3. **Adaptive Reasoning:** 
   - Base your logic strictly on tool outputs and provided context. 
   - If a previous attempt or search fails, you MUST analyze the failure and use a fundamentally different approach or keywords. 
   - Never repeat the exact same tool call or parameters.
4. **Self-Correction & Reflection Loop:** After you propose an <answer>, a "System Reflection Prompt" will ask you to evaluate it. 
   - Output <PASS> and a reason if the answer is fully verified.
   - Output <REJECT> and a reason if the answer is flawed, inconsistent, or lacks evidence.
   - **Continuous Solving:** If you output <REJECT>, you must immediately resume the standard solving process. Your very next response must start with <thinking> to analyze the rejection reason and execute a new tool call to explore a better solution. Continue this "Think-Act-Reflect" cycle until a verified answer is produced.

## Output Format (strict):
Always start with <thinking>. Then, depending on the case, output one of the following:

1. If reasoning continues: 
<thinking> Identify what is known and what is missing. Explain why the following tool call is the most effective next step to resolve the query. </thinking> 
<tool_call> {"name": "...", "arguments": {...}} </tool_call>

2. If ready to conclude: 
<thinking> Summarize the key evidence gathered through your reasoning process and derive the answer clearly. </thinking> 
<answer> Final answer </answer>
\end{lstlisting}

\paragraph{Inspector 1 Prompt.}
This prompt is inserted after Solver 1 proposes an answer.
\begin{lstlisting}[
    basicstyle=\footnotesize\ttfamily,
    breaklines=true,
    breakatwhitespace=false,
    columns=fullflexible,
    keepspaces=true,
    numbers=none,
    frame=single,
    caption={Inspector 1 reflection prompt.}
]
System Reflection Prompt: System Reflection: You must act as a strict reasoning auditor. 
Examine your previous `<answer>` and the entire search trajectory. Does the answer perfectly and factually address the user's original question based ONLY on the gathered evidence?
- If YES: Output `<PASS>` followed by a brief reason explaining why the evidence supports it.
- If NO: Output `<REJECT>` followed by a brief reason explaining the exact logical flaw, missing evidence, or hallucination.
Do NOT plan the next steps here; just provide the label and the reason (For example, using tool call).
\end{lstlisting}

\paragraph{Solver 2 Prompt.}
This bridge prompt is inserted when Inspector 1 rejects the previous answer and control returns to the solver.
\begin{lstlisting}[
    basicstyle=\footnotesize\ttfamily,
    breaklines=true,
    breakatwhitespace=false,
    columns=fullflexible,
    keepspaces=true,
    numbers=none,
    frame=single,
    caption={Solver 2 continuation prompt.}
]
You have rejected the previous answer. Please resume the standard solving process. Start with <thinking> to analyze your rejection reason and explore a new tool call to find the correct answer.
\end{lstlisting}

\end{document}